\documentclass[format=acmsmall, review=false, screen=true,nonacm]{acmart}

\usepackage{amsmath}
\usepackage{booktabs}
\usepackage{graphicx}
\graphicspath{{Figures}}
\DeclareGraphicsExtensions{.pdf,.png}
\usepackage{bigstrut}
\usepackage{multirow}
\usepackage{multicol}
\usepackage{enumitem}
\usepackage{algorithm, setspace}
\usepackage{algpseudocode}
\usepackage{csquotes}
\usepackage{balance}
\usepackage{bm}
\usepackage{ulem}

\newcommand{\p}[1]{\smallskip \noindent \textbf{{#1}.}}
\newcommand{\argmax}{\mathop{\arg\max}}

\newcommand{\eq}[1]{Equation~(\ref{eq:#1})}
\newcommand{\fig}[1]{Figure~\ref{fig:#1}}

\begin{document}

\title{Safe Interactions via Monte Carlo Linear-Quadratic Games}


\author{Benjamin A. Christie}
\orcid{0000-0003-3785-0319}
\email{benc00@vt.edu}

\author{Dylan P. Losey}
\orcid{0000-0002-8787-5293}
\email{losey@vt.edu}

\affiliation{
  \institution{Virginia Tech}
  \department{Department of Mechanical Engineering}
  \streetaddress{635 Prices Fork Rd}
  \city{Blacksburg}
  \state{VA}
  \postcode{24060}
  \country{USA}
  }

\thanks{This work is supported in part by NSF Grant \#2129201.}
 
\begin{abstract}
Safety is critical during human-robot interaction.
But --- because people are inherently unpredictable --- it is often difficult for robots to plan safe behaviors.
Instead of relying on our ability to anticipate humans, here we identify robot policies that are robust to unexpected human decisions.
We achieve this by formulating human-robot interaction as a zero-sum game, where (in the worst case) the human's actions directly conflict with the robot's objective.
Solving for the upper value of this game provides robot policies that {combine} safety and performance across a wide range of human actions.
Existing approaches attempt to find these optimal policies by leveraging Hamilton-Jacobi analysis (which is often intractable) or linear-quadratic approximations (which are often inexact).
By contrast, in this work we propose a computationally efficient and theoretically supported method that approaches a robust, Stackelberg-style policy. 
Our approach (which we call \textit{MCLQ}) leverages linear-quadratic games to obtain an initial guess at safe robot behavior, and then iteratively refines that guess with a Monte Carlo search.
Not only does MCLQ provide real-time safety adjustments, but it also enables the designer to tune how conservative the robot is --- preventing the system from focusing on unrealistic human behaviors.
Our simulations and user study suggest that this approach advances safety for human-robot interaction in terms of both computation time and expected performance.
\end{abstract}

%
%




\maketitle


\section{Introduction}

\begin{figure}[t]
    \centering
    \includegraphics[width=0.6\linewidth]{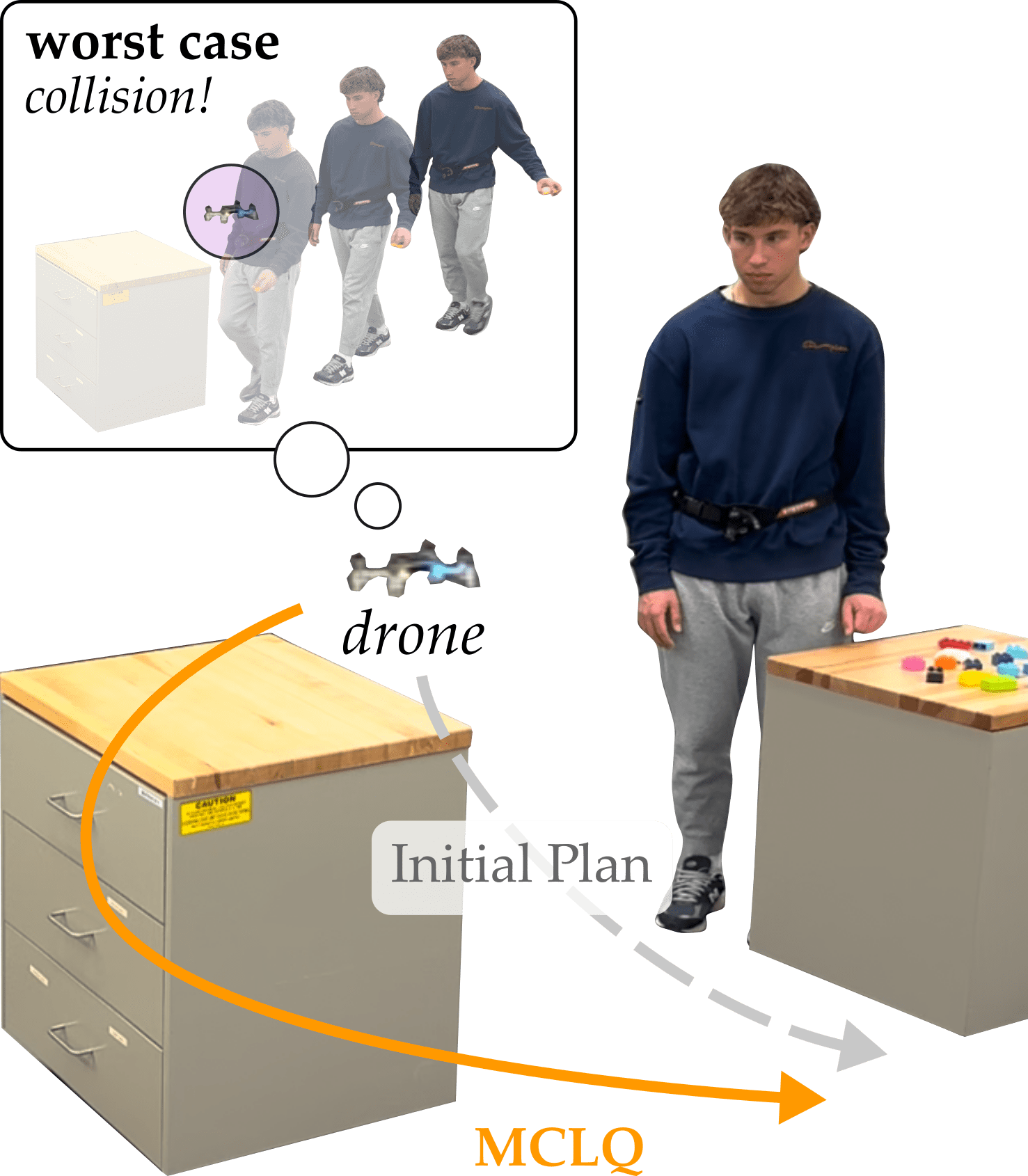}
    \caption{Human and drone moving in a shared workspace. Under our proposed {method (MCLQ)}, the drone reasons about worst case human actions within designer-specified bounds, and then selects safe behaviors in response to those actions. For instance, here the drone moves across the table to prevent a potential collision.}
    \label{fig:front}
\end{figure}

Interacting with people is challenging because humans are inherently unpredictable.
Consider \fig{front} where an autonomous drone is flying near a human worker. 
This robot has some high-level task it wants to complete (e.g., environment monitoring), as well as a low-level controller which dictates how the robot should accomplish this task (e.g., circling the room). 
If the robot knew precisely what the human was going to do, it could anticipate the human's actions and choose behaviors that maintain a safe distance between agents.
But because people often take unexpected actions, real human behavior will deviate from the robot's model.
As a result, the robot's original plan --- which it thought was safe --- may actually be unsafe. 
Returning to our example, a drone that turns left (because it predicts the human will stop) actually puts both agents in danger (because the human unexpectedly keeps walking forwards).

To address this problem, today's robots recognize that they are often uncertain about the human's actions. 
Instead of assuming the human will follow an exact model, these robots search for plans that are safe across a distribution of human behaviors. 
There are two general approaches here.
The first is a \textit{precise} method: the robot reasons over the distribution of possible human trajectories, and then chooses the optimal control policy that maximizes safety across the distribution \cite{bansal2017hamilton, hsu2023safety, fisac2019bridging}.
The second approach is based on \textit{approximations}: the robot simplifies the system dynamics and objectives, and then obtains a closed-form policy that maximizes safety under the simplified setting \cite{tian2022safety, fridovich2020efficient}.
Unfortunately, both types of approaches have significant limitations.
Theoretically exact solutions are often computationally intractable; e.g., our autonomous vehicle could not apply these methods in real-time to adjust its behaviors.
On the other hand, approximate solutions are imprecise, {and may result in unsafe behaviors that are \textit{locally} optimal but globally unsafe as the system becomes increasingly complex and nonlinear.}


In this paper we propose an {action-augmentation method} for human-robot interaction that robots can leverage to tune their real-time behavior and enhance safety around unpredictable humans. 
Our proposed {method} achieves precise and tractable performance by combining both types of prior approaches. 
Specifically, our insight is that:
\begin{center}
\textit{We can leverage approximate methods to seed safe robot behavior, and then apply \\a stochastic local search to guide that initial guess towards optimal solutions.}
\end{center}
We show an example of our method in \fig{front}.
The robot approximates the interaction as a linear-quadratic (LQ) system, and finds an action sequence that maximizes robot performance under the worst case human response.
In parallel, the robot refines this action sequence using sampling methods that noisily converge towards the {open-loop optima} of a zero-sum game {while stochastically escaping local optima}.
Robots that execute the resulting actions are robust to unpredictable humans within designer-specified bounds: even if the person suddenly changes direction, the autonomous drone has planned a safe trajectory.

Overall, we make the following contributions:

\p{Identifying Robust Behavior}
To maintain safety despite unpredictable humans, we frame interaction as a zero-sum game. 
Here the robot tries to minimize its cost while realizing an adversarial human might take actions to maximize that cost.
The game's upper value defines a robot policy that is robust to unexpected human behavior; we develop a method (MCLQ) to tractably approximate this robust policy.

\p{Uniting Prior Works} 
Our approach to find robust robot policies blends aspects of approximate and precise methods.
Specifically, we leverage LQ games to get an initial action sequence, and then refine that sequence with Metropolis-Hastings sampling. 
Our analysis proves that in the best case MCLQ converges to the Nash Equilibrium, and in the worst case our proposed approach is as good as current approximations.

\p{Adjusting Safety Margins}
Our MCLQ framework enables practical advantages that go beyond existing methods. 
For example, designers can constrain the safety margin, i.e., the range of human actions the robot reasons over.
Decreasing this safety margin causes the robot to rely more on its human predictions, preventing the robot from focusing on unrealistic human behaviors and becoming overly conservative.

\p{Conducting Experiments} 
We conduct simulations and a user study to compare our approach to state-of-the-art methods. 
In simulation, we show that MCLQ solves for safe control policies more rapidly than HJB and iLQ methods, while also achieving higher performance than existing iLQ approximations.
In the real world, participants reported feeling ``safer'' when interacting with drones that leverage MCLQ compared to alternatives, due to MCLQ producing more ``predictable'' and ``attentive'' behavior. MCLQ is able to complete tasks more quickly and safely than iLQ alternatives.

\section{Related Work}

Related research has introduced a variety of control algorithms to ensure safety during human-robot interaction \cite{hsu2023safety}.
Within this larger field, our approach is most connected to game-theoretic controllers.
These methods frame the human and robot as two agents with their own objectives, and solve for robot policies that are robust (i.e., minimize worst-case costs) while accounting for the human's decisions.

\p{Exact Methods}
Game-theoretic works treat human-robot interaction as interconnected system: both agents have a task they are trying to accomplish, and the actions of each agent affect one another.
We can formalize these interconnected dynamical games with Hamilton-Jacobi equations \cite{bacsar1998dynamic, wabersich2023data}.
Precisely solving the Hamilton-Jacobi equations provides the optimal, game-theoretic behavior for each agent; e.g., a policy for the autonomous car that avoids collisions with the human.
Hamilton-Jacobi analysis has therefore become a gold standard for safe interaction \cite{bansal2017hamilton, bansal2020hamilton, fridovich2020confidence, bajcsy2019efficient, bansal2021deepreach}.
Unfortunately, most Hamilton-Jacobi equations have no analytical solution, and numerical methods are often intractable --- requiring both exponential time and computational memory \cite{starr1969nonzero, fridovich2020efficient}.
As such, within this paper we develop a real-time approximation that converges towards the ideal solution provided by the Hamilton-Jacobi framework.

\p{Linear-Quadratic Approximations}
We are not the first to try to approximate the game-theoretic solution.
For example, \cite{bansal2021deepreach, jiang2016using, nakamura2021adaptive, nakamura2020causality, fisac2019bridging} propose neural network models, and \cite{fridovich2020efficient, jones2020polynomial, chilan2020optimal} outline open-loop methods. 
One particularly promising approximation recognizes that we can analytically solve the Hamilton-Jacobi equations in a special case: if the system dynamics are linear and the costs are quadratic \cite{bacsar1998dynamic}.
For these linear-quadratic (LQ) games the optimal robot policy is closely connected to linear-quadratic regulators \cite{wu2023learning, an2023fuzzy}.
Robots can solve LQ games quickly, providing a tractable method to obtain safe behaviors that match the gold standard of Hamilton-Jacobi analysis.
Indeed --- even if the system is not linear-quadratic --- recent works have tried to apply sequential linear-quadratic (iLQ) approximations \cite{fridovich2020efficient, laine2023computation}.
Under these iLQ approaches the robot iteratively linearizes the dynamics and quadraticizes the cost before solving the resulting LQ game to find the human and robot control gains.
On the one hand, iLQ provides an increasingly prevalent method to tractably identify safe controllers.
But on the other hand, this approximation is fundamentally limited by its reliance on linear-quadratic systems --- as the real interaction diverges from the simplification, iLQ falls short.
Overall, this gap motivates our work towards methods that combine theoretical exactness and practical implementation.

\p{Accounting for the Human}
Using the game-theoretic solution leads to policies that are robust to worst-case disturbances.
However, doing so can make robot policies overly conservative --- even when the human is not interfering with the robot's objective. 
Recent works \cite{sagheb2025unified, tian2022safety, hu2023deception, hu2024active, fridovich2020efficient, nam2024active} have attempted to mitigate this issue by using a model of the human that updates over the course of an interaction.
The authors of \cite{tian2022safety} maintain an estimate of the current user's \textit{role} in a Stackleberg-style formulation. 
If they are a follower, then the robot must be more cautious, whereas if they are a leader, then the robot can find calculate an optimal response.
Unfortunately, this approach is limited to settings where a backwards-reachable tube (representing the agents' actions) can be calculated in real-time, such as in linear-quadratic games.
Likewise in \cite{hu2024active}, a shielding-aware dual control method is proposed that has theoretical guarantees on robot safety while maintaining an estimate of the uncertain agent's parameters. This sampling-based approach is similar to our proposed approach, but it is limited to control-affine dynamics and Boltzmann-rational human models.
Other works such as \cite{nam2024active, kedia2024interact, parekh2023learning, christie2024limit, parekh2025using} leverage learned models that adapt to the human and predict their future intent. Although these approaches effectively optimize for performance in the \textit{average} interaction, they are not robust to \textit{worst-case} actions that the human {may} select. 
In this work we propose a flexible, real-time approach that accounts for worst-case human actions and determines robot responses which are robust to these actions within designer specified bounds.

\section{Problem Statement}\label{sec:problem}

We are interested in interactions between a single human and a single robot.
The robot has a task to perform, and to complete this task the robot must reason over the human (e.g., avoid colliding with the human worker).
Without loss of generality, we assume the robot has an initial control policy for the task, as well as a nominal model for predicting the human's behavior.
{Our approach will adjust the robot's policy to enhance safety even when the human deviates from the nominal model. To achieve this safety, we consider the worst-case interaction (within designer-specified bounds): i.e., we identify which actions the human can take that will have the worst impact on the robot's performance.}
Below we formalize this problem setting.
%

\p{Zero-Sum Game}
Let the system have state $x \in \mathcal{X}$ (e.g., the position of both agents).
The robot takes action $u \in \mathcal{U}$, and the human takes action $w \in \mathcal{W}$ (e.g., the human and robot velocities).
At each timestep $t$, the system state transitions according to the deterministic, discrete-time dynamics:
\begin{equation}
    x^{t+1} = f(x^t, u^t, w^t)
\end{equation}
An interaction lasts for a total of $T$ timesteps. 
The system begins the interaction in state $x^0$ and during the interaction it follows a trajectory $\xi = \{(x, u, w)^0, \cdots, (x, u, w)^{T-1}, x^T\}$. 
Since the dynamics are deterministic, we can abbreviate this trajectory as a sequence of robot and human actions $\xi(x^0) = \{(u, w)^0, \cdots, (u, w)^{T-1}\} = (u^{0:T}, w^{0:T})$. 

The robot's goal is to select actions that will cause the system state to update in a way that completes the robot's task.
More formally, the robot has a cost function it seeks to \textit{minimize} across the interaction:
\begin{equation}
\begin{gathered}
    J\left(x^t, u^{t:T}, w^{t:T}\right) =
    \sum\limits_{\tau = t}^{T - 1} j(x^\tau, u^\tau, w^\tau)
  + D(x^T)
\\
    \text{s.t.}~x^{\tau+1} = f\left(x^\tau, u^\tau, w^\tau\right)
\end{gathered}
\label{eq:zero-sum-cost}
\end{equation}
Here $J$ is the cumulative cost, $j(x, u, w)$ is the cost at a single timestep, and $D(x^T)$ is the bequest state cost.
We emphasize that the robot's cost in \eq{zero-sum-cost} depends on the human's actions $w$. 
For example, the autonomous drone will incur a significant penalty if the human moves into that agent.
To optimize cost, the robot has an initial policy $\hat{\pi}_\mathcal{R}$ that determines how it will complete the task.
The robot may also have some guess for how the human agent will behave: $\hat{\pi}_\mathcal{H}$.
Both policies are mappings from states $x$ to actions:
\begin{align}
\hat{\pi}_\mathcal{R}&: \mathcal{X} \mapsto \mathcal{U}
    \\
\hat{\pi}_\mathcal{H}&: \mathcal{X} \mapsto \mathcal{W}
\end{align}
In practice, the real human will inevitably deviate from the robot's model.
A worst-case human selects actions that \textit{maximize} the cost in \eq{zero-sum-cost}.
Formally, in this worst-case the human and robot are participating in a two-player \textit{zero-sum} game \cite{bacsar1998dynamic}: the robot is trying to minimize its cost, whereas the antagonistic human is attempting to maximize that same cost, directly opposing the robot's objective.
One benefit of focusing on the worst-case human is that --- if the robot's behavior is safe in this worst case --- we know it is safe for other human responses.

In this work, we assume that the cost function $J$ and state transition function $f$ are sufficiently smooth and have continuous first- and second-derivatives with respect to $x$. We further assume that there exists a norm $\| \circ \| $ such that $\| \nabla_x f(x, u, w)\| \le 1$, i.e. the system state is controllable and does not grow unbounded for finite actions. Additionally, we assume that the action sets $\mathcal{U}$ and $\mathcal{W}$ are compact, bounded, and convex. In practice, this assumption is naturally satisfied by the limits of robotic constraints and human biomechanics. This assumption is not required by our proposed method, but it is necessary for the following robust minimax formulation.

\p{Robust Minimax Formulation and Game Values}
Similar to prior works \cite{tian2022safety, riskawareslumbers2023game, wang2020game}, we have formulated human-robot interaction as a zero-sum game. The advantage of this formulation is that identifying robot behavior that is safe in the worst-case ensures safety regardless of the human's actual deviation.
To determine what this worst-case behavior is, we turn to the bounds of optimal behavior in zero-sum games.
In general, zero-sum games are characterized by their upper and lower values. The lower value of the game represents a scenario where the human must commit to an action sequence first, allowing the robot to optimally respond: 
\begin{equation}
    \uline{V}(x) = 
    \sup\limits_{w^{0:T} \in \mathcal{W}^T}
    \inf\limits_{u^{0:T} \in \mathcal{U}^T}
    \left(
    J(x, u^{0:T}, w^{0:T})
    \right)
    \label{eq:lower-value}
\end{equation}
In this situation, the robot is at an advantage. 
Note that the inner $\inf$ operator is evaluated for a specific, fixed human action $w^{0:T}$.
Since the robot can see what the human's action is, it can always choose the best response that minimizes the overall cost $J$.
Conversely, the upper value of the game represents the scenario where the robot commits to a trajectory, and the adversarial human responds with the worst-case optimal sequence:
\begin{equation}
    \overline{V}(x) =
    \inf\limits_{u^{0:T} \in \mathcal{U}^T}
    \sup\limits_{w^{0:T} \in \mathcal{W}^T}
    \left(
    J(x, u^{0:T}, w^{0:T})
    \right)
\end{equation}
Here, the robot is at a disadvantage. It cannot observe the human's action: instead, it must plan an action that minimizes the overall cost, regardless of what the human chooses. 
Since the human --- wanting to maximize the robot's cost --- can observe the robot's action, they can always choose an action that leads to worst-case performance.
A pure-strategy equilibrium point between these two values only exists if the Isaacs' Condition holds (for details, see Sion's Minimax Theorem in \cite{sion1958general} and Isaacs' Condition in \cite{isaacs1999differential}).
However, because our interaction involves complex non-linear dynamics (such as articulated manipulators) and non-quadratic objective costs (such as logarithmic barrier functions), the cost landscape $J$ is generally neither strictly convex in $u$ nor strictly concave in $w$, which violates Isaacs' condition. Thus, we recognize that there may be an information gap between the two values where $\overline{V}(x) \ge \uline{V}(x)$.

To guarantee safety, it is strictly required that the robot prepares for the worst-case: the \textit{upper} value of the game. By planning against $\overline{V}(x)$, the robot identifies a robust Stackelberg-style policy where it acts as the leader, committing to an open-loop plan and anticipating the worst-case best-response from the human follower. Therefore, our objective is to find the robust open-loop robot policy such that:
\begin{equation}
\pi_\mathcal{R} = 
\mathop{\arg\inf}\limits_{u^{0:T} \in \mathcal{U}^{T}}
\sup\limits_{w^{0:T} \in \mathcal{W}^{T}}
\left(
J(x^t, u^{0:T}, w^{0:T})
\right)
\label{eq:robot-policy}
\end{equation}
In other words, the robot policy should consider the worst-case action of the human before it selects its own action. The robot policy that maximizes \textit{robustness} follows the upper value of the game.

\p{Intractability of the Robust Minimax}
Directly solving for the robust policy $\pi_\mathcal{R}$ is computationally intractable for continuous, high-dimensional action spaces. While traditional approaches attempt to solve the continuous-time counterpart via the Hamilton-Jacobi-Isaacs (HJI) partial differential equations to find global feedback policies, such methods suffer heavily from the curse of dimensionality and cannot be run in real-time. 
Instead, other approaches aim to find a \textit{locally} robust policy by approximating the system as linear-quadratic.
The information gap that results from this approximation is theoretically $0$ and an optimal robot policy can be found in closed form. 
Unfortunately, since the resulting lower and upper values of this approximation are not the same as the original game, the policy that is maximally robust in the approximate game may not be robust in the original system.
Given that the robot has some initial policy $\hat{\pi}_\mathcal{R}$, our goal is to develop a real-time approach that \textit{adjusts} the robot's policy so that it converges towards the upper value of the underlying zero-sum game.
Because we cannot tractably identify this ideal policy through \eq{robot-policy}, we must develop an approximation that the robot can leverage efficiently in real-time.

\section{Monte Carlo Linear-Quadratic Games}\label{sec:method}

In this section we propose our real-time {method for safe interactions} based on zero-sum games. To approximate $\pi_\mathcal{R}$ in real-time, we propose a two-stage approach: (1) an analytical "warm-start" using a Linear-Quadratic (LQ) game approximation, and (2) a dual-loop Metropolis-Hastings search that explores the non-convex landscape to minimize the upper value $\overline{V}$.
Our overall approach --- Monte Carlo Linear-Quadratic Games (\textbf{MCLQ}) --- combines rapid LQ solutions with a parallel local search. We theoretically demonstrate that MCLQ improves upon existing LQ baselines and converges towards a robust robot policy.
We conclude with the practical advantages of our MCLQ framework, including a designer-specified safety margin that prevents the robot from becoming overly conservative (Section~\ref{sec:practice}).
An implementation of MCLQ is available \href{https://github.com/safe-interactions-mclq/mclq}{here}.

%
\subsection{Obtaining an Initial Action Trajectory} \label{sec:lq}

Our method attempts to find the sequence of actions $u^{0:T}$ that optimize \eq{robot-policy}.
To get an initial estimate of these actions, we simplify the actual system into an LQ approximation.
Starting with $f$, we linearize the dynamics:
\begin{equation}
    f(x, u, w, t) 
    \approx 
    A x(t) + B u(t) + D w(t)
    \label{eq:linear}
\end{equation}
and quadraticize the state-action cost function $J$:
\begin{equation}
    J(x, u, w, t)
    \approx
    x(t)^{\prime} Q x(t) + u(t)^{\prime} R_u u(t) - w(t)^{\prime} R_w w(t)
    \label{eq:quad}
\end{equation}
where $\circ^\prime$ represents transposition and $Q, R_u, R_w \succ 0$.
In this localized, quadratic regime, the cost function is strictly convex in $u$ and strictly concave in $w$. Consequently, the Isaacs' condition holds locally, and a unique feedback Nash Equilibrium exists.
{We perform this LQ approximation about the most recently calculated trajectory. If this approximation is being performed at $t=0$, then we sample a random inital trajectory}. 
Note that this LQ approximation is processed for the entire trajectory across timesteps $\hat{t} \in \{t, \ldots, T + t\}$.
Put another way, the LQ approximation is performed for each state-action pair in the trajectory.
Under this LQ approximation the Nash Equilibrium policies exhibit linear state-feedback behavior \cite{bacsar1998dynamic}.
These policies are captured by the gain matrices:
\begin{equation}
\begin{split}
    u(t) &= -K(t) x(t)
    \\
    w(t) &= -L(t) x(t)
\end{split}
\end{equation}
Crucially, we do not use these gains for closed-loop control. Instead, we use them to generate an initial open-loop seed by rolling out the linearized dynamics. This seed provides an informed starting point, ensuring that the subsequent stochastic search begins in a region of high dynamic feasability. 
To find the gain matrices, we employ the discrete-time algebraic Riccati Equation (DARE).
For a particular robot gain matrix $K$, the optimal human gain matrix $L$ can be found in closed form by solving the DARE: 
\begin{equation}
\begin{split}
    \bm{P}_{\bm{K}, \bm{L}(\bm{K})} &= \bm{Q} + \bm{K}^\prime \bm{R}_u \bm{K} +
    (\bm{A} - \bm{B}\bm{K})^\prime 
    \widetilde{\bm{P}}_{\bm{K}, \bm{L}(\bm{K})}
    (\bm{A} - \bm{B}\bm{K})
    \\
    \text{s.t.}\quad&
    \Re\left(\|\bm{P}_{\bm{K}, \bm{L}(\bm{K})}\|\right) \ge 0 
    \\
    &\Re\left(\|\bm{R}_w - \bm{D}^\prime \bm{P}_{\bm{K}, \bm{L}(\bm{K})}\bm{D}\|\right) > 0
\end{split}
\label{eq:dare}
\end{equation}
where $\widetilde{\bm{P}}_{\bm{K}, \bm{L}(\bm{K})}$ is condensed for brevity:
\begin{equation}
    \widetilde{\bm{P}}_{\bm{K}, \bm{L}(\bm{K})}
    \equiv
    \bm{P}_{\bm{K}, \bm{L}(\bm{K})}
    +
    \bm{P}_{\bm{K}, \bm{L}(\bm{K})}
    \bm{D}
    \cdot (\bm{R}_w - \bm{D}^\prime \bm{P}_{\bm{K}, \bm{L}(\bm{K})} \bm{D})^{-1} \bm{D}^\prime\bm{P}_{\bm{K}, \bm{L}(\bm{K})}
\end{equation}
and the terminal condition is $\bm{P}_{\bm{K}, \bm{L}(\bm{K})} = \bm{Q}$. 
The bolded matrices shown in \eq{dare} are the compacted notation presented in \cite{wu2023learning} with appropriate padding:
\begin{equation*}
\begin{gathered}
\bm{Q} \equiv \text{diag}({Q^0, \ldots, Q^T}), 
\quad 
\bm{R_u} \equiv \text{diag}(R_u^0, \ldots, R_u^{T-1}),
\quad
\bm{R_w} \equiv \text{diag}(R_w^0, \ldots, R_w^{T-1})
\\
\bm{A} \equiv \begin{bmatrix}
\bm{0} & \bm{0}
\\
\text{diag}({A^0, \ldots, A^T}) & \bm{0}
\end{bmatrix},
\quad
\bm{B} \equiv \begin{bmatrix}
\bm{0}
\\
\text{diag}({B^0, \ldots, B^{T - 1}})
\end{bmatrix},
\quad
\bm{D} \equiv \begin{bmatrix}
\bm{0}
\\
\text{diag}({D^0, \ldots, D^{T - 1}})
\end{bmatrix}
\\
\bm{K} \equiv \begin{bmatrix}
\text{diag}({K^0, \ldots, K^{T - 1}}) & \bm{0}  
\end{bmatrix},
\quad
\bm{L} \equiv \begin{bmatrix}
\text{diag}({L^0, \ldots, L^{T - 1}}) & \bm{0}  
\end{bmatrix}
\end{gathered}
\end{equation*}
The unique human gain matrix $\bm{L}(\bm{K})$ that \textit{maximizes} the quadraticized cost is:
\begin{equation} \label{eq:L}
    \bm{L}(\bm{K}) = (-\bm{R}_w + \bm{D}^\prime \bm{P}_{\bm{K}, \bm{L}(\bm{K})} \bm{D})^{-1} \bm{D}^\prime 
    \bm{P}_{\bm{K}, \bm{L}(\bm{K})}(\bm{A} - \bm{B}\bm{K})
\end{equation}
Note that $\bm{L}(\bm{K})$ is the \textit{worst-case} gain matrix given the robot's policy and objective; we have found the human actions that maximize the robot's cost.
To develop the correct response to this worst-case, we use a similar equation to compute the robot gain matrix $\bm{K}(\bm{L})$:
\begin{equation} \label{eq:K}
    \bm{K}(\bm{L}) = 
    (\bm{R}_u + \bm{B}^\prime \bm{P}_{\bm{K}(\bm{L}), \bm{L}} \bm{B})^{-1}  \bm{B}^{\prime}
 \bm{P}_{\bm{K}(\bm{L}), \bm{L}}\left(\bm{A} - \bm{D}\bm{L}\right)
\end{equation}
{By approximating the system as LQ and recomputing the coupled DAREs for both \eq{L} and \eq{K}, we converge to feedback policies that form the local Nash Equilibrium for zero-sum LQ games.} 
Expanding these matrices provides our initial guess of the optimal action trajectory $\xi(x^0) = (u^{0:T}, w^{0:T})$.
As a reminder, we do not use this trajectory for closed-loop control. Instead, we use this trajectory as a seed for the stochastic search presented in Section~\ref{sec:mc}.

The traditional formulation of linear-quadratic approximations for zero-sum games can violate the eigenvalue conditions of \eq{dare}, even when a closed-form NE exists for the underlying non-LQ system. In practice, this can be alleviated by adjusting $R_u$ and $R_w$ manually.
The gain matrices can be solved directly through \eq{dare} or with readily available open-source software, such as Scipy's \texttt{solve\_discrete\_are}
\cite{virtanen2020scipy}.

\subsection{Refining the Action Trajectory} \label{sec:mc}

The initial action trajectory produced by the local equilibrium policies in Section~\ref{sec:lq} are optimal if the system is linear-quadratic.
However, as established in Section~\ref{sec:problem}, the environments that we are interested in may feature non-linear dynamics and non-quadratic costs, meaning that the LQ solution is merely a local approximation.
These nonlinearities and nonconvexities cause the information gap $\overline{V}(x) - \uline{V}(x) > 0$.
To ensure safety, we must bridge the gap between this local optimum and the true upper value of the game $\overline{V}(x)$.

To tractably solve this nested optimization, we propose using the LQ solution as a highly-informed "warm start" that is subsequently refined through stochastic gradient descent. Specifically, we apply a nested Metropolis-Hastings (MH) sampling algorithm. While other Monte Carlo methods could be substituted within our general framework, the ergodic properties of the MH sampler are particularly well suited for escaping the local optima that frequently trap pure LQ-approximate methods \cite{andrieu2006ergodicity}.

As a reminder, our ultimate goal is to reach the robust robot policy defined in \eq{robot-policy}, which can be treated as a nested optimization problem.
In the outer loop the robot proposes a sequence of actions $u^{0:T}$ to minimize its cost, and in the inner loop the antagonistic human responds with actions $w^{0:T}$ that maximize that cost.
Because the robust policy takes the form of a minmax optimization, our MH sampler is explicitly divided into two coupled phases \cite{hoegerman2023reward}: an inner loop that models the worst-case human response, and an outer loop that optimizes the robot's robust commitment. 
An outline of our resulting approach is shown in Algorithm~\ref{alg:mclq}.


\begin{algorithm}[t]
\caption{Monte Carlo Linear-Quadratic Games (MCLQ)}\label{alg:mclq}
    \begin{algorithmic}[1]
    \Require $f, J, \beta, M, N$ \Comment{Dynamics and Cost Functions}
    \Procedure{LQ Approximation}{$x$, $f$, $J$}
        \State $\bm{A}, \bm{B}, \bm{D} \gets \text{Linearize}(f)$
        \State $\bm{Q}, \bm{R}_u, \bm{R}_w \gets \text{Quadraticize}(J)$
        \State $\bm{K}, \bm{L} \gets \text{DARE Solution}$
        \State $\xi_u \gets -\bm{K} \bm{x}$
        \State $\xi_w \gets -\bm{L} \bm{x}$
        \State \textbf{return} $\xi_u$, $\xi_w$
    \EndProcedure \Comment{Provides initial guess of $\xi_u$, $\xi_w$}
    \Procedure{Monte Carlo Search}{$x$, $\xi_u$, $\xi_w$, $f$, $J$}
        \State $\Delta_u \gets 0$, $\Delta_w \gets 0$
        \State $J_0 \gets J(x, \xi_u, \xi_w)$
        \For{$m = 1 \ldots M$} 
            \Comment{Outer Loop: find the best-case robot action}
            \For{$n = 1 \ldots N$} 
            \Comment{Inner Loop: find the worst-case human response}
                \State $\Delta_n \gets \text{Perturb}(\xi_w, \Delta_w)$
                \Comment{Optionally use the safety margin $\lambda$ or model $\hat{\pi}_H$}
                \State $J_w \gets J(x, \xi_u + \Delta_u, \xi_w + \Delta_n)$
                \If{$J_w > J_0 \lor \exp(\beta (J_w - J_0)) > \eta, ~ \eta \sim U[0, 1]$}
                    \State $J_0 \gets J_w$
                    \State $\Delta_w \gets \Delta_n$
                \EndIf
                 \Comment{Updates worst-case human actions}
            \EndFor
            \State $\Delta_m \gets \text{Perturb}(\xi_u, \Delta_u)$
            \State $J_u \gets J(x, \xi_u + \Delta_m, \xi_w + \Delta_w)$
            \If{$J_u < J_0 \lor \exp(\beta(J_0 - J_u)) > \eta, ~ \eta \sim U[0, 1]$}
                \State $J_0 \gets J_u$
                \State $\Delta_u \gets \Delta_m$
            \EndIf \Comment{Updates robot response to human}
        \EndFor
        \State \textbf{return} $\xi_u' = \xi_u + \Delta_u$, $\xi_w' = \xi_w + \Delta_w$
    \EndProcedure
    \end{algorithmic}
\end{algorithm}

\p{Inner Loop}
The inner loop seeks to identify worst-case human actions that increase the robot's cost.
Given a fixed robot commitment $\xi_u$, the adversarial human wants to maximize $J$. We define the target distribution for the human's trajectory as a Gibbs measure:
\begin{equation}
    P_{\pi_\mathcal{H}}\left(\xi_w \mid \xi_u, x\right)
    \propto
    \exp\Big(\beta J\left(x, \xi_u, \xi_w\right)\Big)
    \label{eq:m1}
\end{equation}
where $\beta > 0$ is the inverse temperature. As $\beta \to \infty$, the exponential function heavily exaggerates differences in $J$ and the probability measure collapses into a Dirac delta function centered strictly on the global maximum of $J$. 
Therefore, sampling from $\pi_\mathcal{H}$ at a high $\beta$ is theoretically equivalent to solving the inner $\sup$ problem.
Sampling directly from this distribution is intractable for continuous and high-dimensional systems. Instead, we use the Metropolis-Hastings sampler \cite{chib1995understanding} by proposing random perturbations $\Delta \xi_w$ of the human's action.
These perturbations are accepted if they increase the total cost, or if they satisfy the acceptance rate:
\begin{equation}
    \exp\left(\beta \cdot \Big(
    J\left(x, \xi_u, \xi_w + \Delta \xi_w\right)
    - 
    J\left(x, \xi_u, \xi_w\right)
    \Big)
    \right) > \eta, \quad \eta \sim U[0, 1]
    \label{eq:inner-acceptance}
\end{equation}
In practice, using a $\beta$ that is too large leads to numerical instability and reduces exploration, so we use a $\beta < 20$ and opt to do the comparison logarithmically.
The inner loop repeats $N$ times before terminating with a new human action sequence $\xi_w' = \xi_w + \Delta \xi_w^N$.
Sampling random perturbations $\Delta \xi_w$ according to this scheme will find human actions that are the worst-case for a given robot action $\xi_u$. 
The mechanism for proposing actions human actions can instead use the nominal human model $\hat{\pi}_\mathcal{H}$; this process is detailed in Section~\ref{sec:practice}.

\p{Outer Loop}
The outer loop takes the updated choice of human actions $\xi_w'$, and seeks to find robot actions that are robust to the human.
Similar to the inner loop, we define the target distribution for the robust robot trajectory as:
\begin{equation}
    P_{\pi_\mathcal{R}}(\xi_u \mid x)
    \propto
    \exp\left(
    -\beta
    \mathop{\mathbb{E}}\limits_{\xi_w \sim P_{\pi_\mathcal{H}}}
    \Big[
    J(x, \xi_u, \xi_w)
    \Big]
    \right)
    \label{eq:m2}
\end{equation}
Like \eq{m1}, directly sampling from this distribution is difficult. We follow the same tractable sampling scheme as before. The outer loop modifies $\xi_u$ by proposing random perturbations $\Delta \xi_u$.
These perturbations are accepted if they decrease total cost, or if they satisfy the acceptance rate:
\begin{equation}
    \exp\left(\beta \cdot \Big(
    J\left(x, \xi_u, \xi'_w\right)
    - 
    J\left(x, \xi_u + \Delta \xi_u, \xi_w'\right)
    \Big)
    \right) > \eta, \quad \eta \sim U[0, 1]
    \label{eq:outer-acceptance}
\end{equation}
Importantly, the inner loop (i.e., the human) is \textit{maximizing} the cost, while here the outer loop (i.e., the robot) is \textit{minimizing} that same cost.
This outer loop terminates after $M$ iterations, and outputs a final robot action trajectory $\xi_u' = \xi_u + \Delta\xi_u^M$.

\subsection{Balancing Performance and Safety} \label{sec:practice}

One key advantage of our approach is that --- as we will show --- it can be implemented in real-time.
But another core aspect is that the designer can tune the \textit{safety margin}, i.e., how conservative the robot's behavior is.
Recall that $\hat{\pi}_\mathcal{H}$ is a predictive model of the human's actions.
We recognize that the real human will inevitably deviate from $\hat{\pi}_\mathcal{H}$; but how different should we expect the human's actions to be?
Within a zero-sum game formulation, if the human can take any action at any time, then the robot is forced to overly conservative behaviors (e.g., the autonomous drone always moving away from the human).
Unlike LQ approximations, our approach is structured to resolve this conflict between performance and safety.
Specifically, during Monte Carlo sampling we can impose hard constraints on the perturbations $\Delta\xi_w$ so that they are close to the predictions of our human model.
Let $\hat{\xi}_w \sim \hat{\pi}_\mathcal{H}$ be a predicted human trajectory. We can limit the range of considered human actions such that:
\begin{equation} \label{eq:dkl}
    \| \hat{\xi}_w - (\xi_w + \Delta\xi_w)\|^2_2 \leq \lambda
\end{equation}
where $\lambda$ is a design parameter that determines conservatism. 
As $\lambda \to \infty$, the search space approaches the original trajectory space $\Xi_\mathcal{H}$, and as $\lambda \to 0$, the robot is increasingly confident in its human model.
Importantly, \eq{dkl} does not determine \textit{if} the human's behavior aligns with our nominal model: it restricts the search space to trajectories that are \textit{similar} to our nominal model.

Instead of restricting the sampled actions such that they are close to the nominal model's prediction, we can change the sampling distribution of \eq{m1} entirely. If we want the adversarial inner-loop to align with our expectation of the human's behavior, we can constrain the Gibbs measure of \eq{m1} to be:
\begin{equation}
\argmax\limits_{P \in \mathcal{P}_{\pi_\mathcal{H}}}
~D(P ~\|~ \hat{\pi}_\mathcal{H}),
\quad \text{s.t.} \; D(P ~\|~ \hat{\pi}_\mathcal{H}) \le \lambda
\end{equation}
where $D$ is a divergence metric (such as the Kullback-Leibler Divergence \cite{kullback1951information}). This accomplishes the same goal as \eq{dkl}: limiting conservatism in the robot policy. However, instead of enforcing the restriction $\lambda$ on the sampled trajectories, we instead enforce it on the action distribution itself.
In practice, this is accomplished by changing the perturbation distribution $\Delta\xi_w \sim P$ from a uniform distribution to one that aligns more closely --- according to the conservatism parameter $\lambda$ --- with the nominal human model.

\subsection{Summary}\label{sec:method-summary}

Taken together, the LQ game in Section~\ref{sec:lq} provides a first guess at the robot's action trajectory $\xi$.
This action trajectory is then refined by nested Metropolis-Hastings samplers in Section~\ref{sec:mc} that update $\xi$ to solve for the robust robot policy defined in \eq{robot-policy}.
The resulting trajectory $\xi'$ contains the sequence of actions a robust robot should execute over the next $T$ timesteps; this robust action sequence {is} recomputed online (i.e., at each timestep $t$). Notably, we compute the LQ game approximation \textit{once} at the beginning of the interaction and use the nested Metropolis-Hastings samplers thereafter.
To limit conservatism, the inner loop of the nested Metropolis-Hastings samplers can be tuned with a design parameter $\lambda$ and a nominal human model using Section~\ref{sec:practice}.
In what follows, we will first theoretically justify that this approach is strictly better than alternatives that just use a local LQ-approximation, then experimentally validate this claim in Sections~\ref{sec:sims} and \ref{sec:us}.

The Metropolis-Hastings samplers presented in Section~\ref{sec:mc} asymptotically converge to the target distributions $P_{\pi_\mathcal{R}}$ and $P_{\pi_\mathcal{H}}$. Since we run each sampler for a finite time, the samplers are not guaranteed to produce optimal trajectories. 
Following from \cite{chib1995understanding, tierney1994markov}, the probability of a specific optimality gap $r \ge 0$ is approximately on the order of:
\begin{equation}
\begin{gathered}
P(\| \xi^\prime - \xi^\star \| \le r)
\gtrsim 
1 - \mathcal{O}(e^{-\beta \Delta(r)}) - \mathcal{O}(e^{-MN})
\\
\Delta(r) = \inf
\left[
J\left(\xi_u^\prime, \xi_w^\star\right)
- J\left(\xi_u^\star, \xi_w^\prime\right)
\right]
\end{gathered}
\label{eq:bounds}
\end{equation}
after $MN$ iterations of the nested Metropolis-Hastings samplers, where $\xi^\star$ is the optimal trajectory.
If the cost function is convex in $\xi_u$ and concave in $\xi_w$ (e.g., if the system has quadratic cost and separable dynamics), then the relation simplifies to:
\begin{equation}
\begin{gathered}
    P\left(\| 
    \xi^\prime - \xi^\star \|
    \le r\right)
    \gtrsim 
    1 - 2 e^{-\beta r^2 / 2} - \mathcal{O}(e^{-MN})
\end{gathered}    
\label{eq:conv-bounds}
\end{equation}
With sufficient $N$, $M$, and $\beta$, MCLQ will approach the stationary distributions in Equations~(\ref{eq:m1}) and (\ref{eq:m2}) and therefore produce the robust robot policy shown in \eq{robot-policy}.
Now, we consider three cases for the underlying environment: if the dynamics are linear and the cost is quadratic, if the dynamics are nonlinear, and if the system is non-linear-quadratic.

\begin{itemize}
    \item
    \textbf{LQ Systems:}
    For a linear-quadratic system the DARE presented in \eq{dare} finds the equilibrium point between the two agents, if it exists.
    This means that --- for our MCLQ method --- the initial trajectory is the most robust robot policy, and Monte Carlo sampling is unable to improve upon this initial guess. 
    \item
    \textbf{Non-Linear Systems with Quadratic Cost:}
    If the system is non-linear but has quadratic cost, then MCLQ improves upon the original LQ approximation \textit{proportional to the error in the dynamics linearization}. 
    The resulting error follows \eq{conv-bounds}. As an example, for a separable $4$-dimensional system evaluated over a control horizon of $10$ steps, the probability that the trajectory produced by MCLQ is within one unit of the optimal trajectory approaches $99\%$ with $\beta = 10$ as the total number of iterations in the stochastic search grows.
    \item 
    \textbf{Non-Linear Non-Quadratic Systems:}
    If the system is not linear-quadratic, then MCLQ improves upon the LQ approximation with a probability shown in \eq{bounds}. If the error in the LQ approximation is large, then the sampler will improve upon the initial trajectory with probability proportional to $1 - \exp(-\beta \Delta(r))$.
\end{itemize}
\section{Simulations}\label{sec:sims}

\begin{figure}
\centering
    \includegraphics[width=0.5\linewidth]{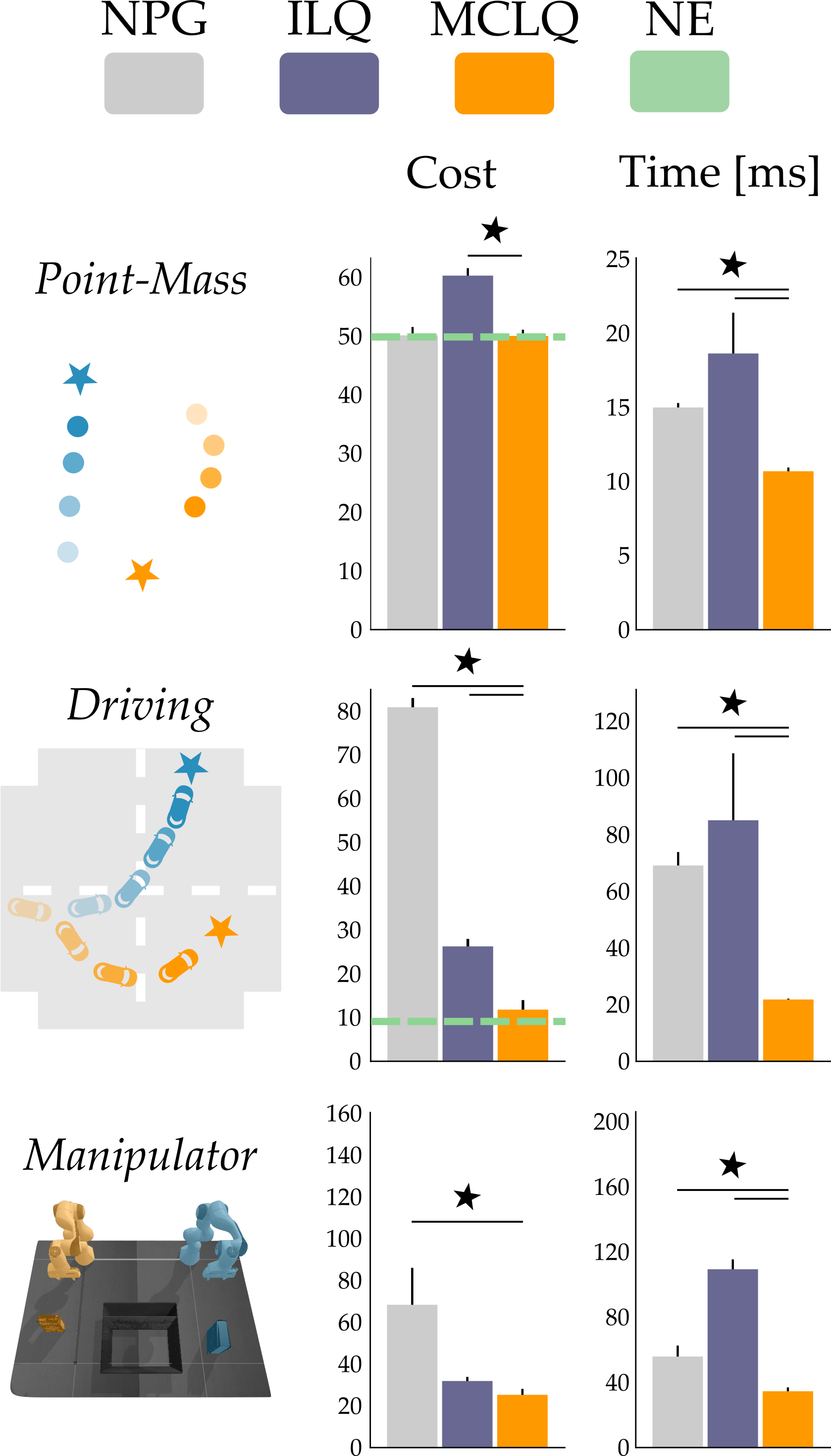}
    \caption{Simulation results across \textit{point-mass}, \textit{driving}, and \textit{manipulator} environments. (Left) We plot the cost and (Right) computation time averaged over $100$ simulations. 
    Computation time is the number of milliseconds per robot action (normalized by the number of timesteps per trajectory). 
    In non-LQ settings the computation time for {NE} is prohibitively high; e.g., in \textit{driving} the {NE} computation time exceeded one hour. We did not calculate {NE} in the $26$-dimensional \textit{manipulator} environment. 
    Error bars show standard deviation and an $*$ denotes statistical significance.}
    \label{fig:sims1}
\end{figure}

In theory, our proposed MCLQ method moves towards robust, game-theoretic behaviors while offering real-time performance.
Here we test our theoretical claims by comparing our approach to exact Hamilton-Jacobi methods and tractable LQ approximations.
More specifically, we perform controlled experiments where simulated agents interact in three environments: point-mass, driving, and manipulator (see \fig{sims1}).
Each environment contains two agents.
The simulated robot is attempting to complete its task (e.g., reaching a goal) while simulated humans move in its proximity.
The robot must avoid collisions with these simulated humans --- even when the humans take unexpected or noisy actions.

\p{Independent Variables} 
We vary the robot's controller across four levels.
To test LQ approximations, we include \textbf{DARE-nA} \cite{wu2023learning} and \textbf{ILQGames} \cite{fridovich2020efficient} approaches. 
These methods iteratively approximate the dynamics and cost as a LQ system and solve for the resulting Nash Equilibrium. We used the official repositories for \textbf{DARE-nA} and \textbf{ILQGames} when applicable. 
To test exact solutions, we next solve for the true Nash Equilibrium (\textbf{NE}) using the Bellman Equation.
Finally, we evaluate our \textbf{MCLQ} approach from Section~\ref{sec:method}.

\p{Environments} 
Below we describe the three simulated environments.
All simulations were performed on an AMD Ryzen 7000 Series 5 CPU with multithreading enabled.

\smallskip

\noindent\textit{Point-Mass.}
Here the simulated human and robot move in a $2$D plane.
The initial positions and velocities of both agents are randomized; across an interaction of $T=30$ timesteps, the robot attempts to reach a fixed goal location.
Both agents have linear dynamics such that $x^{t+1} = x^t + u^t + w^t$, and the robot's cost is quadratic (considering both distance to goal and distance to human).
Because this environment is a linear-quadratic game, we anticipate that LQ approximations should find the Nash Equilibrium.

\smallskip 

\noindent\textit{Driving.}
Our driving environment is taken from the iLQ baseline \cite{fridovich2020efficient}.
As before, one simulated human and robot move in a $2$D plane with randomized initial configurations.
The dynamics of each agent follow a nonlinear bicycle model. 
Here the robot's cost function is not quadratic:
\begin{equation*}
    j(x, u, w) = \|x_\mathcal{R}^t - g\|^2 + \eta \exp\left(\frac{\|x_\mathcal{R}^t - x_\mathcal{H}^t\|^2}{-\eta}\right)
    + u^\prime R_u u
    \label{eq:cost-driving}
\end{equation*}
where $g$ is the goal position the robot is trying to reach, $x_\mathcal{R}$ is the robot's position, $x_\mathcal{H}$ is the human's location, and $\eta$ scales the cost of approaching the human.
Each interaction lasts a total of $T=30$ timesteps.

\smallskip

\noindent\textit{Manipulator.}
Our final environment contains two $7$-DoF robot arms in PyBullet.
We treat one of these arms as the simulated human, and the other as the robot.
Both agents are trying to pick up and place objects within a shared workspace.
The dynamics of the arms and the task are nonlinear, and the system state $x \in \mathcal{X}$ is $26$-dimensional (including manipulators and objects).
We leverage a cost function similar to \textit{driving}.
The robot moves to reach, grasp, and place items while avoiding collisions with the other arm.
We separate this task into two separate sub-tasks: \textit{reaching} and \textit{placing}, each with separate cost functions. 
Settings with grasping violate the smooth dynamics assumptions of Section~\ref{sec:problem}. 
To still use LQ approximations for the environment, the sub-task is automatically switched when the \textit{reaching} sub-task's cost is below a certain threshold.

\p{Simulated Human}
We simulate humans as bounded-rational agents that noisily optimize their cost function $J_H$:
\begin{equation}
\begin{split}
    \pi_\mathcal{R}\left(w^t \mid x^t\right) 
    &\propto 
    \exp\left(-\alpha \cdot
    J_H(x^t, w_i^t)
    \right)
    \\
    \text{s.t.}~x^{t+1} &=f(x^t, u^{t_0}, w^t)
\end{split}
\label{eq:human-model-br}
\end{equation}
Increasing $\alpha \rightarrow \infty$ causes the human to always take the optimal action, while decreasing $\alpha \rightarrow 0$ causes the human to act randomly \cite{fridovich2020confidence}.
In our simulations we set $\alpha = 7.5$.
When computing the cost for future timesteps the simulated human assumes the robot will repeat its most recent action $u^{t_0}$ (e.g., the robot will keep moving with the same velocity).
In each environment the human had a task that was independent of the robot's objective --- for instance, in \textit{driving} the human tried to reach their own goal position.

\p{Results}
The results from our first simulation are summarized in \fig{sims1}. 
Across all three environments, {MCLQ} achieved costs that were closest to the upper value of the game (\textbf{NE}) while also requiring the least amount of computation time.
In \textit{point-mass} we highlight that all methods reached similar cost; this matches our expectations, because \textit{point-mass} was an LQ system.
By contrast, in non-LQ systems (\textit{driving} and \textit{manipulator}) the LQ approximations made by {{DARE-nA}} and {ILQ} fell short, leading to suboptimal performance.
Computing the exact solution with {NE} was time-consuming and not always feasible: indeed, in the $26$-dimension \textit{manipulator} task, we were unable to compute the true {NE} because of the high-dimensional and continuous state-action space.

\p{Adjusting Safety Margins}
As discussed in Section~\ref{sec:practice}, one feature of our approach is the designer-selected safety margin.
MCLQ robots reason over the worst-case human action within bounds $\lambda$.
By increasing $\lambda$ in \eq{dkl} the designer makes the robot more risk-averse (i.e., the robot considers larger deviations from the nominal human model).
Conversely, decreasing $\lambda$ makes the robot more risk-seeking (i.e., the robot increasingly relies on its human model).
Within our simulations from \fig{sims1} we left this value of $\lambda$ fixed at risk-neutral behavior.
Now we explore how increasing and decreasing $\lambda$ changes the robot's performance in a modified \textit{point-mass} environment.

\begin{figure}
    \centering
    \includegraphics[width=1.0\linewidth]{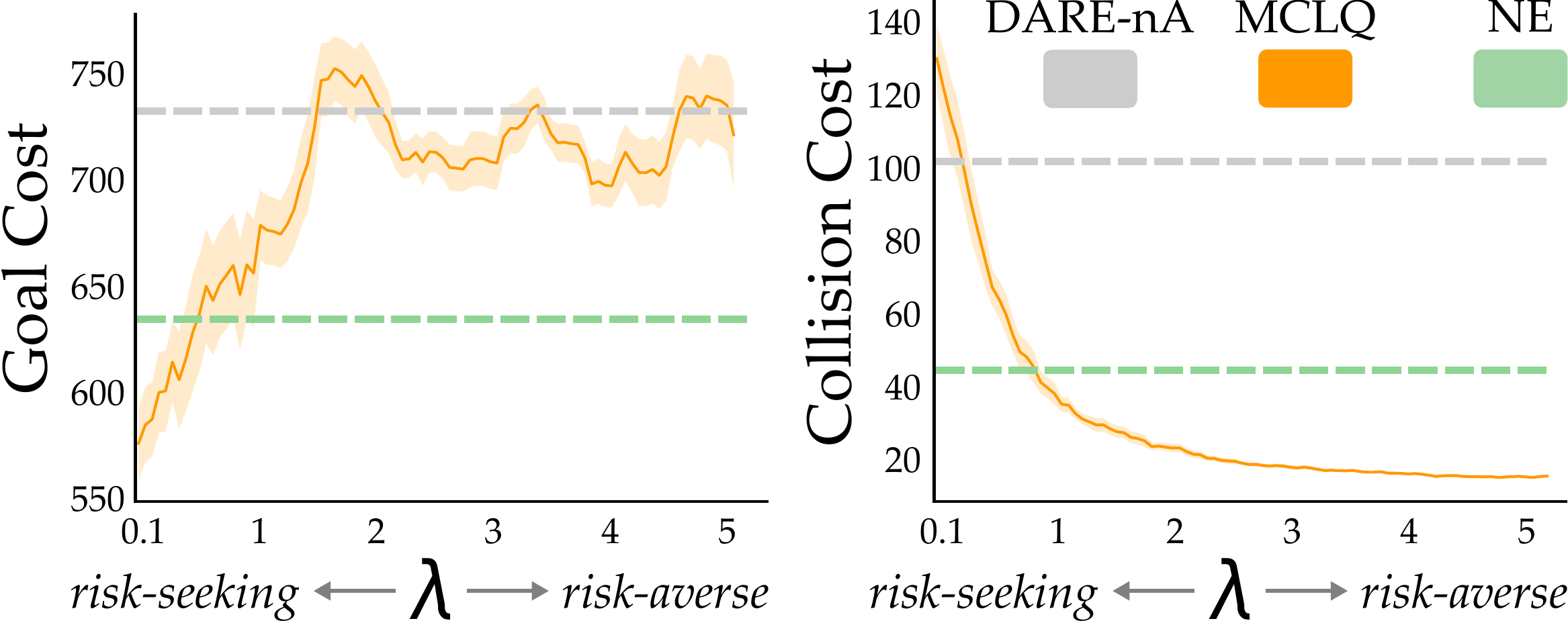}
    \caption{
    Simulation results for a modified point-mass environment where we adjust the safety margin $\lambda$ in MCLQ. Increasing $\lambda$ causes the MCLQ robot to consider a wider range of worst case human actions, resulting in more conservative behavior. Conversely, decreasing $\lambda$ causes the MCLQ robot to increasingly rely on its nominal human model. Unlike LQ approximations, our proposed method gives designers the flexibility to tune $\lambda$ and adjust the safety margin. In practice, this results in risk-averse robots that avoid all possible collisions but reach the goal more slowly, or risk-seeking robots that allow some potential collisions and rapidly move to the goal.}
    \label{fig:sims2}
\end{figure}

Our extended \textit{point-mass} environment includes {one to ten humans} that are navigating to goal positions.
The robot seeks to reach to its own goal while avoiding these randomly generated agents.
To find an initial plan, the robot is equipped with a nominal human model --- it assumes each human will move in a straight line towards their goal.
In practice, however, our simulated humans deviate from this model when nosily optimizing their cost function. \fig{sims2} plots the robot's performance as a function of $\lambda$. 
When $\lambda$ increases the robot becomes more risk-averse: the MH sampler accounts for a wider range of adversarial human actions. 
This causes the robot agent to stay farther from humans (reducing collisions), but also leads to longer paths (increasing distance). 
Conversely, lower values of $\lambda$ cause to risk-seeking behavior where the robot relies on its human model. 
If humans stick to this model, the robot avoids collisions while minimizing distance.
But when humans deviate, the number of collisions increase.
Overall, this simulation supports our theoretical description of $\lambda$ and demonstrates how designers can leverage MCLQ to tune the safety margin.

We further test how designers can change the human model within Equations~(\ref{eq:m1}) and (\ref{eq:inner-acceptance}) by comparing the performance of MCLQ across different human models.
We leverage the noisly-optimal human model from \eq{human-model-br} with both varying levels of rationality $\alpha \in \{2.5, 10.0, 20.0\}$ and cost functions $J_H \in \{J_H^1, J_H^2, J_H^3\}$:
\begin{align}
J_H^1(x, u, w) &= \|x_H^N - g_H\|^2_{P_H} + \sum\limits_{t = 0}^{N-1} \|x_H^t - g_H\|^2_{Q_H} + \|w^t\|^2_{R_H}
\label{eq:e1}
\\
J_H^2(x, u, w) &= \|x_H^N - g_H\|^2_{P_H} + \sum\limits_{t = 0}^{N-1} \|x_H^t - g_H\|^2_{Q_H} + \eta \exp\left(\frac{\|x^t_R - x^t_H\|^2_2}{-\eta}\right) + \|w^t\|^2_{R_H} 
\label{eq:e2}
\\
J_H^3(x, u, w) &= \|x_H^N - g_H\|^2_{P_H} + \sum\limits_{t = 0}^{N -1} \|x_H^t - x_R^t\|^2_{Q_H} + \|w^t\|^2_{R_H}
\label{eq:e3}
\\
\text{s.t.} \quad  x^{t + 1} &= f(x^t, u^t, w^t), \quad u^t \sim \pi_\mathcal{R}(x^t)
\end{align}
where $Q_H$, $R_H$, and $P_H$ are fixed, time-invariant matrices and $\pi_\mathcal{R}$ is our proposed method for selecting robust robot actions. These cost functions cause the human to move towards the goal while minimizing their action size, but they also ensure that the human follows one of three general behaviors: ignore the robot's position (\ref{eq:e1}), avoid the robot's position (\ref{eq:e2}), and follow the robot (\ref{eq:e3}).
Intuitively, each of these cost functions could be a reasonable approach for predicting the human's behavior, and designers may choose any of these (or other similar functions).
Our goal is to demonstrate that even when the human model is incorrect --- and the human follows another underlying behavior pattern --- our approach to robust interact still provides a safety measure.

\begin{figure}
    \centering
    \includegraphics[width=0.5\linewidth]{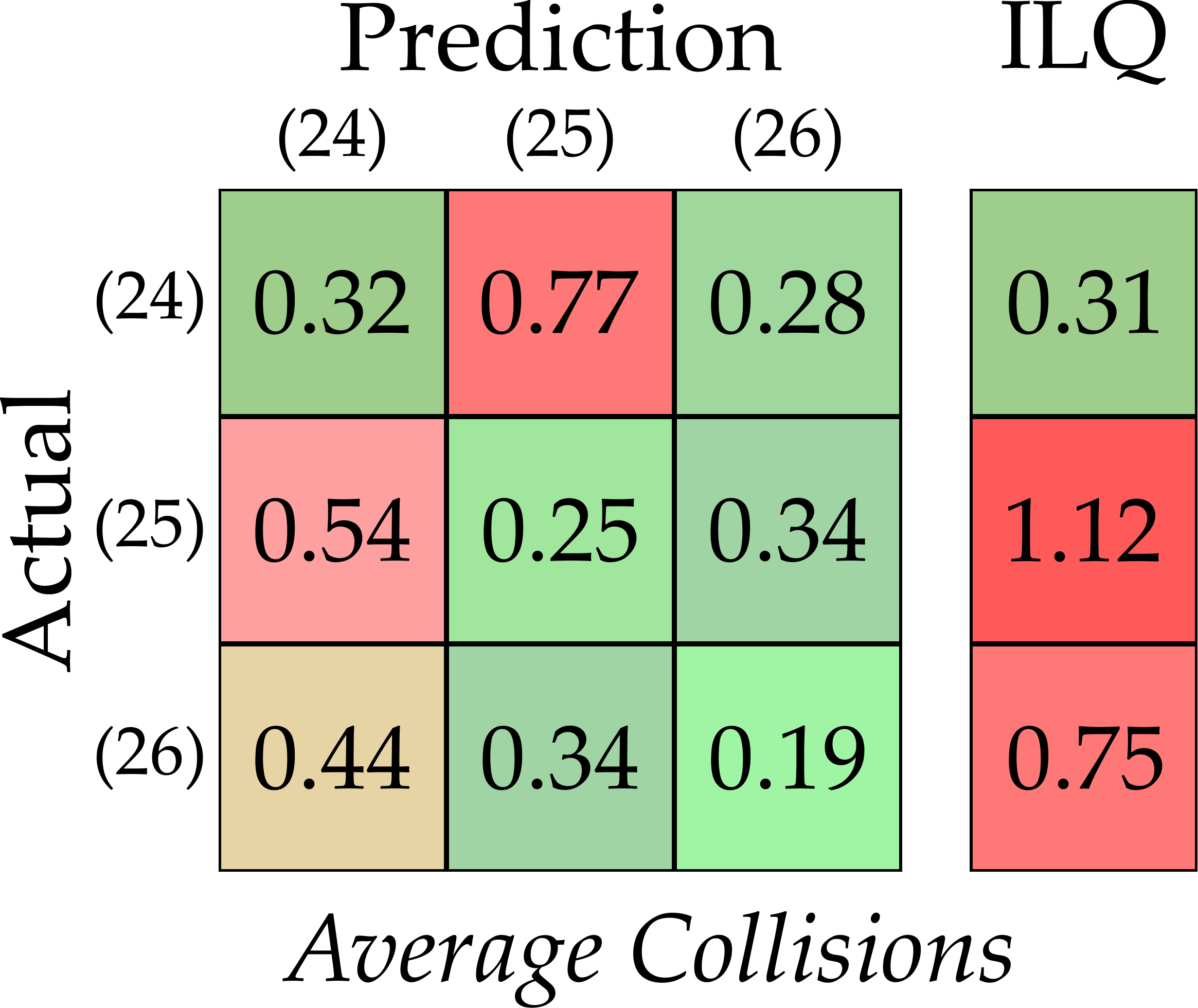}
    \caption{Generally, when the human model aligns with the actual human behavior, MCLQ avoids worst-case scenarios (here, collisions).
    The annotations (24), (25), and (26) correspond to the human models in Equations~(\ref{eq:e1}-\ref{eq:e3}).
    Here we define a collision as an instance where the human is within $5$ units of the robot. The results are averaged across $1089$ trials with $9$ variations of human models in the point-mass environment.
    Note that values along the diagonal tend to be the lowest, indicating that when the actual and modeled motions of the human model align, the robot maintains safety. 
    Overall, when a pessimistic human model is used (such as \eq{e3}), the robot tends to avoid the human more frequently.
    In this experiment, we define a ``collision'' as a state where the human and robot agents are within $5$ units of each other. The initial configuration of the agents is randomly sampled from $U[-100, 100]$. Similar to \textit{Point Mass}, each agent has two degrees of freedom (i.e., $\mathcal{X} = \mathbb{R}^4$ and $\mathcal{U} = \mathcal{W} = \mathbb{R}^2$) and each interaction lasts for $T = 100$ timesteps.
    }
    \label{fig:human-models}
\end{figure}

As expected, when the actual human aligns with the behavior that MCLQ is modeling, the number of average collisions decreases (\fig{human-models}).
Notably, when the predicted human model is more pessimistic (such as in \eq{e3}), MCLQ tends to avoid the human more at the expense of goal convergence. 
We test MCLQ with a safety margin $\lambda = 1.0$ according to \eq{dkl} and a non-quadratic cost function similar to the \textit{Driving} environment. 
Notably, the iterative LQ approximation baseline has similar performance for \eq{e1}, but fails for complex behaviors that are more difficult to approximate. 

\section{User Study}\label{sec:us}

Our simulations from Section~\ref{sec:sims} support our theoretical analysis, and suggest that MCLQ converges towards maximally-robust behaviors while minimizing computation time.
We next evaluate our method in a real-world setting with $N=24$ in-person human participants.
Here users walk around a room to complete an assembly task while an autonomous drone circles that room to inspect the parts (see \fig{front}).
The robot modifies its high-level trajectory to avoid getting to close to the human workers.
We compare two \textit{real-time} {methods for safe interactions that adjust} the drone's behavior: {ILQ} \cite{fridovich2020efficient} and {MCLQ}.
We selected {ILQ} here because it was the best performing real-time baseline from our simulations.

\p{Experimental Setup}
Participants interacted with a Crazyflie 2.1+ (Bitcraze) during the assembly task.
The participant's objective was to construct a Lego tower at the central station.
The blocks needed for building that tower were scattered in three other stations located along the perimeter of the workspace.
Accordingly, users needed to move back and forth through the workspace to acquire blocks and build their tower.
The drone's objective was to monitor the workspace during the task by completing as many revolutions around the central station as possible.
As the drone moved around the central station it repeatedly intersected with the human worker: here the drone should take actions that avoid the moving participant.
{Participants interacted with the drone separately (i.e., the drone interacted with one user at a time).}

\p{Participants and Procedure}
We recruited $24$ participants ($6$ female, age $24.5 \pm 4.5$) from our university community for this user study. 
Of the $24$ participants, $8$ had not used drones and $5$ did not have experience with robotics. 
Participants received monetary compensation for their time and provided informed written consent according to university guidelines (IRB \#23-1237).

We leveraged a between-subjects design where every participant interacted with {ILQ} for five minutes and {MCLQ} for five minutes.
To prevent the participants from always going back and forth between the same stations, we instructed users to move according to three movement patterns: clockwise, counter-clockwise, and random.
Both the order of the methods and movement patterns were counterbalanced (i.e., half of the participants started with {MCLQ}).
Participants were never told which algorithm the drone was using.

\p{Dependent Measures --- Objective}
We using tracking devices (VIVE) to measure the states and actions of the drone and human at each timestep of the interaction.
To assess objective performance, we considered two metrics.
For our first metric (\textit{collisions}) we counted the number of times the drone was within a radius of $0.5$m from the human. 
Our second metric (\textit{revolutions}) is the number of revolutions the drone completed within the five minute trial. 
Lower values for \textit{collisions} indicate that the drone is maintaining safety, while higher values for \textit{revolutions} show that the drone is performing the task more efficiently.

\p{Dependent Measures --- Subjective}
After interacting with each control algorithm, participants completed a $7$-point Likert scale survey. 
This survey assessed the user's subjective preferences along three multi-item scales. We asked users:
\begin{enumerate}
    \item If they felt \textit{safe} during the interaction,
    \item If they thought the drone was \textit{attentive} to their position,
    \item If they thought the drone's movements were \textit{predictable}.
\end{enumerate}

\p{Hypotheses}
We had two hypotheses for this user study:
\begin{quote}
   \textbf{H1.}~\textbf{MCLQ} \textit{will modify the drone's actions to reduce the number of collisions and improve subjective feelings of safety}.
\end{quote}
\begin{quote}
   \textbf{H2.}~\textbf{MCLQ} \textit{will complete a similar number of revolutions as the iLQ baseline}. 
\end{quote}

\begin{figure*}
    \centering
    \includegraphics[width=1.0\linewidth]{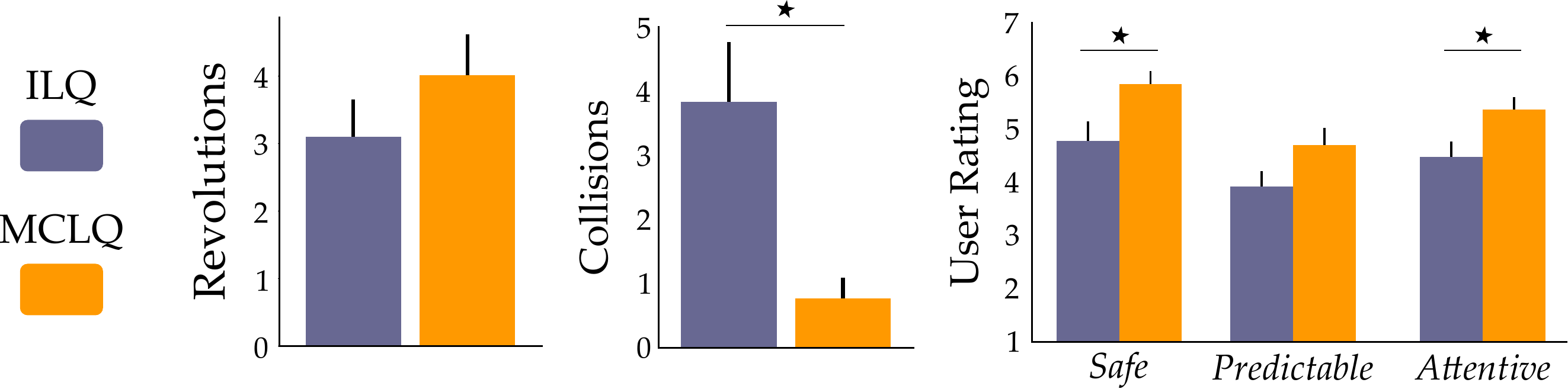}
    \caption{Results from our user study in Section~\ref{sec:us}. Participants walked around a room to assemble a tower; a drone completed revolutions around the same workspace to monitor the human's progress (also see \fig{front}). (Left) The average number of revolutions the drone completed and the average number of collisions. Here ``collisions'' occurred when the drone was within $0.5$ meters of the human. The proposed {MCLQ} algorithm adjusts the robot's behavior to increase safety (fewer collisions) while also enhancing performance (more revolutions). (Right) After interacting with each algorithm participants answered survey questions about how safe, predictable, and attentive the robot was. Ratings suggest that participants perceived {MCLQ} to be a safer system. Error bars show standard deviation and an $*$ denotes statistical significance ($p < .05$).}
    \label{fig:us}
\end{figure*}

\p{Results}
The results from our in-person user study are summarized in \fig{us}.
To assess \textbf{H2}, we measured the number of revolutions that each method completed per interaction: a higher number indicates more efficient performance. 
One-way ANOVA tests show that the performance difference between the methods is trending towards significance ($F(2, 86) = 1.50$, $p = 0.14$), where {MCLQ} completes more revolutions than {ILQ}.
This suggests that the safety improvements made by {MCLQ} are not coming at the expense of overly conservative behavior --- the {MCLQ} drone is still performing its high-level task of monitoring the workspace. 

Given that the drone is completing a similar number of revolutions with each method, the key question becomes the \textit{safety} of the robot's behavior.
We analyzed \textbf{H1} along two levels: objective safety (maintaining a minimum distance between agents) and subjective safety (the user's perception of the system).
For objective safety, ANOVA tests revealed that {MCLQ} leads to significantly fewer collisions than {ILQ} ($F(2, 86) = 4.16$, $p < 0.001$).
The participants' responses to our Likert scale survey align with these objective results: users perceived the {MCLQ} robot to be safer ($F(2, 24) = 2.33$, $p < 0.05$) and more attentive ($F(2, 24) = 2.37$, $p < 0.05$).
In addition, users thought that drones following the {MCLQ} algorithm had more predictable actions, with differences trending towards significance ($F(2, 24) = 1.84$, $p = 0.073$).
In their free response comments, participants stated that with the {MCLQ} drone they ``\textit{felt safer}'' and that the drone was ``\textit{more reactive and predictable}.''
\section{Conclusion} \label{sec:conclusion}

In this paper we presented a real-time safety approach for human-robot interaction.
Our approach ensures that the robot is robust to noisy and unexpected human behaviors within designer-specified bounds.
To achieve this game-theoretic safety with tractable computation, we combined linear-quadratic approximations with stochastic local searches.
Our theoretical and empirical analysis showed the resulting MCLQ algorithm converges towards {robust robot policies} while providing flexible and efficient implementation.
Across multiple simulations and a user study, we observed that MCLQ advances both objective and subjective safety measures when compared to state-of-the-art control alternatives.

\p{Limitations}
MCLQ is a step towards control policies that are robust to human disturbances. 
Since the stochastic search is performed on the underlying system directly, MCLQ is not restricted to systems that can be approximated as linear-quadratic. 
One key assumption in our work is that the objective function and state dynamics can be evaluated quickly. 
The inner loop of the double Metropolis-Hastings sampler is somewhat expensive: every perturbation of the original trajectory mandates recalculating the dynamics and cost.
Performing this recalculation quickly is not always feasible, such as in fluid dynamics or with visuomotor policies.



\bibliographystyle{ACM-Reference-Format}
\bibliography{bibtex}


\end{document}